# Application of machine learning for hematological diagnosis

Gregor Gunčar[1,¶], Matjaž Kukar[1, ¶], Mateja Notar[1], Miran Brvar[2,3], Peter Černelč[4], Manca Notar[1], Marko Notar[1,*]

[1] Smart Blood Analytics Swiss SA, CH-7000 CHUR, Switzerland
[2] Centre for Clinical Toxicology and Pharmacology, Division of Internal Medicine, University Medical Centre Ljubljana, SI-1000 Ljubljana, Slovenia
[3] Institute of Pathophysiology, Faculty of Medicine, University of Ljubljana, SI-1000 Ljubljana, Slovenia
[4] Department of Hematology, Division of Internal Medicine, University Medical Centre Ljubljana, SI-1000 Ljubljana, Slovenia
* Corresponding author
E-mail: marko@sba-swiss.com
¶ These authors contributed equally to this work.

**Quick and accurate medical diagnosis is crucial for the successful treatment of a disease. Using machine learning algorithms, we have built two models to predict a hematologic disease, based on laboratory blood test results. In one predictive model, we used all available blood test parameters and in the other a reduced set, which is usually measured upon patient admittance. Both models produced good results, with a prediction accuracy of 0.88 and 0.86, when considering the list of five most probable diseases, and 0.59 and 0.57, when considering only the most probable disease. Models did not differ significantly from each other, which indicates that a reduced set of parameters contains a relevant "fingerprint" of a disease, expanding the utility of the model for general practitioner's use and indicating that there is more information in the blood test results than physicians recognize. In the clinical test we showed that the accuracy of our predictive models was on a par with the ability of hematology specialists. Our study is the first to show that a machine learning predictive model based on blood tests alone, can be successfully applied to predict hematologic diseases and could open up unprecedented possibilities in medical diagnosis.**

Machine learning has undergone significant development over the past decade and is already used successfully in many intelligent applications covering a wide array of data related problems[1]. One of the most intriguing questions is whether it can be successfully applied to the field of medical diagnostics and what kind of data is needed? There are several examples of the successful application of machine learning methods in specialized medical fields[2-6], and most recently, a model capable of classifying skin cancer based on images of the skin, with a level of competence comparable to that of a dermatologist has been presented[7]. There are however, no successful applications of machine learning that would tackle the broader and more complex fields of medical diagnosis, such as hematology.

Medical diagnosis is the process of determining which disease explains a person's symptoms and signs. In a physician's ability to arrive at a differential diagnosis and to plan quickly, his or her medical knowledge, skills and experience will play a significant role[8].

In a diagnostic procedure, available information is complemented by additional data gathering, which can be obtained from a patient's medical history, a physical examination and from various diagnostic tests, including clinical laboratory tests. Laboratory tests are used to confirm, exclude, classify or monitor diseases and to guide treatment[9]. However, the true power of laboratory test results is frequently underestimated, since clinical laboratories tend to report test results as individual numerical or categorical values, with physicians concentrating mainly on those values that fall outside a given reference range[10].

The clinical diagnosis of hematological diseases is primarily based on laboratory tests of blood and even the most skilled hematology specialist can overlook patterns, deviations and relations between the increasing numbers of blood parameters being measured in a modern laboratory. Alternatively, machine learning algorithms can easily handle hundreds of attributes (parameters) and are capable of detecting and utilizing the interactions among them, which makes this field of medicine particularly interesting for machine learning applications. Our hypothesis was that the "fingerprint" of a certain hematological disease that we find in blood test results values is sufficient for a machine learning based predictive model to suggest a plausible diagnosis, provided that it has learned to recognize it from a sufficiently large dataset of medical cases, described by clinical laboratory blood tests that are associated with a correct diagnosis determined by a hematology specialist, who has utilized all of the diagnostic procedures necessary to confirm it.

In this study, we describe and evaluate two Smart Blood Analytics (SBA) hematological predictive models based on two different sets of clinical laboratory blood test results (with a different number of blood parameters) and coded diseases, and their evaluation. We evaluated both models in using stratified ten-fold cross-validation of 8233 cases, as well as 20 additional randomly selected hematological cases and compared their performance against an evaluation made by hematology and internal medicine specialists. We also present an illustrative example of how our SBA predictive model can help speed up the diagnostic process.

## Materials and Methods

### Study setting and population

We collected data provided by the Clinical Department of Hematology for patients admitted to the University Medical Centre Ljubljana (UMCL) between 2005 and 2015. The hospital is a tertiary referral center located in Ljubljana, Slovenia and serves a local population of 400,000 inhabitants. For each admitted adult patient (18+ years of age) we collected anonymized laboratory blood test results and their diagnoses made on admission and on discharge. To minimize any bias from previous treatments we only considered a patient's first admittance during the sampling period. In total, we collected data on 8233 cases for which 371341 laboratory blood tests were performed. We then manually curated the data and identified 181 different blood tests that were performed at least 10 times (see Supplementary Table S1 online). On average 24.9% (45 parameters) were measured in every case (Fig. 1).

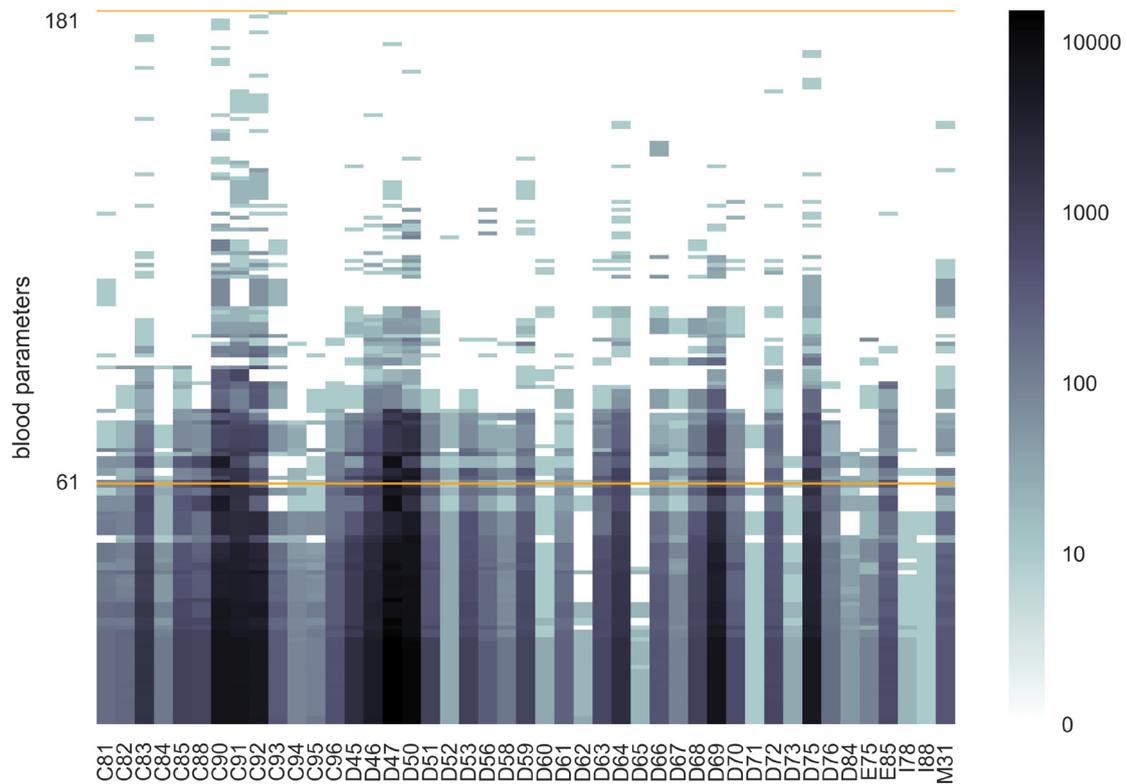

**Figure 1. Parameter coverage for 43 different hematological categories of diseases.** Frequency of every parameter measured is depicted with different colors and marked on the scale bar. The orange line delineates the reduced set of 61 basic parameters from the 181 blood test parameters measured.

From 181 unique blood tests we further selected a reduced subset that contained 61 of the most frequently measured basic blood parameters (see Supplementary Table S1 online). We then excluded all of the specific laboratory tests that are used to confirm a specific diagnosis. On average 66.4% (41 parameters) were measured for every case (Fig. 1).

To compare diagnostic performance of physician and our predictive models, we selected an additional 20 random anonymous cases, which had received their first hematological diagnoses in 2016 or 2017 (see Supplementary Data S2 online). The data included laboratory blood tests from 20 adult patients comprising 10 women and 10 men.

For recording purposes, the UMCL uses a modified International Statistical Classification of Diseases and Related Health Problems (ICD). Our learning models used recorded diseases to three-characters deep in the ICD hierarchy. In total, 43 different hematological categories of diseases were identified among the 8233 cases analyzed (Table 1, Supplementary Table S3 online).

**Table 1. Top ten most prevalent categories of diseases in UMCL, Division for Hematology.**

| ICD code | Disease Category | Frequency | Prevalence |
|---|---|---|---|
| D47 | Other neoplasms of uncertain or unknown behaviour of lymphoid, haematopoietic and related tissue | 1522 | 18.5% |
| D50 | Iron deficiency anaemia | 1190 | 14.5% |
| D69 | Purpura and other haemorrhagic conditions | 743 | 9.0% |
| C90 | Multiple myeloma and malignant plasma cell neoplasms | 739 | 9.0% |
| C91 | Lymphoid leukaemia | 696 | 8.5% |
| C92 | Myeloid leukaemia | 578 | 7.0% |
| D75 | Other diseases of blood and blood-forming organs | 547 | 6.6% |
| D46 | Myelodysplastic syndromes | 457 | 5.6% |
| D64 | Other anaemias | 218 | 2.7% |
| D45 | Polycythaemia vera | 204 | 2.5% |
| Other | | 1339 | 16.1% |

**Predictive model building using a random forest algorithm**

The *random forest* algorithm[11] is a special kind of ensemble approach. Ensembles are a divide-and-conquer approach used to improve predictive performance in machine learning. Ensemble methods[12,13] combine *weak learners* to form a *strong learner*. A random forest consists of many very small decision or regression trees. Each tree, individually, is a weak learner, while all the trees (i.e., a forest) taken together form a strong learner.

Several authors have shown that random forest algorithms perform well in most problem domains[14], including medical diagnostics[15,16]. When compared with hundreds of other machine learning approaches applied to many datasets[17], random forests emerged as the best performer overall.

Random forests are very fast both for training and for prediction, since they can be efficiently parallelized. They perform very well without any parameter tuning, however, they may over-fit particularly noisy data. They are also able to deal with unbalanced and missing data, as well as with huge numbers of attributes and classes (for classification). While our data has low noise (all blood parameters are determined automatically), it is highly dimensional (181 attributes – blood parameters) with many missing values (on average, <25% of parameters were measured for each patient) and has a relatively high number of classes (43), making random forests the logical choice. In our early experiments exploring other popular machine learning methods i.e., decision trees, SVM, and Naive Bayesian classifier, resulted in significantly lower performance scores.

**Smart Blood Analytics algorithm**

The machine learning pipeline (Smart Blood Analytics algorithm) consists of several processing stages:

1. Data acquisition: acquiring raw data from the database
2. Data filtering: selecting only those blood tests performed at the start of the treatment, and the final diagnosis as the machine learning supervisor
3. Data preprocessing: canonizing blood parameters (match with our reference parameter database, filtering erroneous values, handling missing values (imputation)
4. Data modeling: building the predictive model
5. Evaluation: evaluating the predictive model with stratified 10-fold cross-validation)

After having successfully evaluated the data, the predictive model is deployed from the Smart Blood Analytics website (www.smartbloodanalytics.com). If not, the process resumes at one of the earlier stages (data preprocessing or modeling) (Fig. 2).

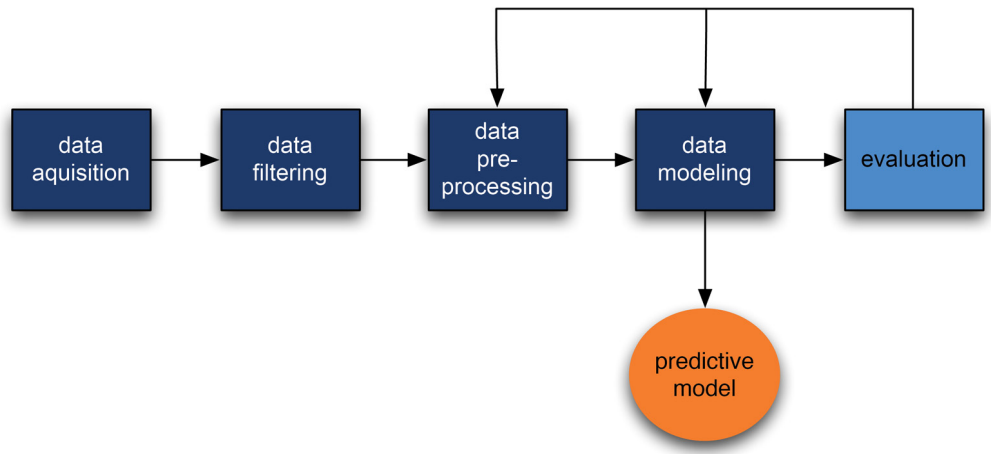

Figure 2. Schematic representation of the Smart Blood Analytics (SBA) algorithm process

**Evaluation of predictive models**

The models were automatically evaluated using stratified tenfold cross-validation, where the folds are selected so that the distribution of diseases is approximately equal in all of the folds. The process was repeated 10 times; each time one fold was set aside for testing, while the remaining 9 folds were used for training. Results were used to make a statistical comparison of the model's performance on all 8233 cases using the Wilcoxon signed-rank test, aggregated into confusion matrices, and performance measures such as diagnostic accuracy, specificity and sensitivity, as well as ROC curves.

**ROC curves**

A common approach for evaluating machine learning results for classification (diagnostic) problems is to observe classification accuracy (true positive rate) obtained by trained classifiers using relevant data sets. However, this approach is valid only under the assumption of uniform error costs (all errors are equally costly). In practice, this is often not the case. For example, in medical diagnostics, a false positive may result in unnecessary health care costs, while a false negative may endanger the patient's life due to delayed treatment. It is therefore often better to observe how the trained classifier behaves in a more general setting.

A popular method for visualizing classifier behavior is by utilizing ROC curves[18]. A ROC curve depicts the relation between the classifier's true positive rate (*sensitivity*) and false positive rate (1–*specificity*). For two-class problems and *scoring* classifiers i.e., a classifier, which produces a score for each possible class, and predicts the class with the maximum score, ROC curves are produced in a straightforward manner by ordering the maximum scores and varying the decision threshold[19]. The one-vs-all approach yields $N$ ROC curves that can be useful in observing classifier performance for each class. Multi-class problems with $N$ classes, can be transformed into $N$ two-class problems, where each problem deals with discriminating one class-vs all other $N-1$ classes. The predictive model's performance is deemed useful for a certain disease (class) if the entire ROC curve is above the diagonal (possibly as close to the upper left corner as possible). ROC curves near the diagonal are of little use, as the predictive model's performance in such cases is only marginally better than chance.

We applied two approaches that are suitable for observing overall classifier performance in multi-class problems: *macro-averaging* and *micro-averaging*[20]. In *macro-averaging*, the ROC curves for all $N$ diseases were averaged regardless of their frequencies. In *micro-averaging* the ROC curve was calculated anew, based upon true positive and false positive rates for all diseases (or equivalently, by weighting ROC curves by relative frequencies of respective diseases and then averaging them).

**Clinical test setting**

We also made two clinical tests, first with five hematology specialists and second with eight non-hematology specialists from internal medicine with at least eight years of experience. Each individual physician in both groups received the laboratory blood test results from 20 cases and was asked to determine a maximum of five potential hematological diseases and to sort them by decreasing order of likelihood. The same 20 cases were then used to make predictions using both predictive models.

**Web-based application and graphical representation of predictive model results**

As part of our work we also developed a web-based application that enables the easy input of blood parameters and produces an innovative representation of the results. It is available upon registration for medical professionals at www.smartbloodanalytics.com. It also includes a novel visualization method of the machine learning model's results (Fig. 3), by showing the 10 most probable diseases depicted as a polar chart with varying radii. Each chart segment represents a possible disease. Each segment's angle corresponds to the predicted (posttest) disease probability. The radius of each segment is proportional to the logarithm of the ratio between the posttest and pretest (prevalence) probability, also called the *information score*. The radius therefore depicts the information (in bits) that is provided by the blood tests that favor (or disfavor) the corresponding disease.

$$r(x) = \log \frac{predicted(x)}{prevalence(x)} = \log predicted(x) - \log prevalence(x) \quad (radius) \quad (1)$$

$$\phi(x) = 2\pi \, predicted(x) \quad\quad\quad (angle\ in\ radians)\ (2)$$

In these two equations (1,2), $r$ is the radius (information score) and $\phi$ the angle (predicted probability) of a disease $x$. Positive radii correspond to information scores in favor of a given diagnosis, while negative radii indicate that blood tests speak against a particular disease (information score is negative). As we cannot depict negative radii, the inner circle in the graph offsets the radius zero (indifferent information). Negative radii is therefore represented inside the innermost circle.

If we look at the graph as a whole, it is strongly related, though not identically, to the Kullback-Leibler[21,22] divergence between posttest (predicted) and pretest (prevalence) probability distributions.

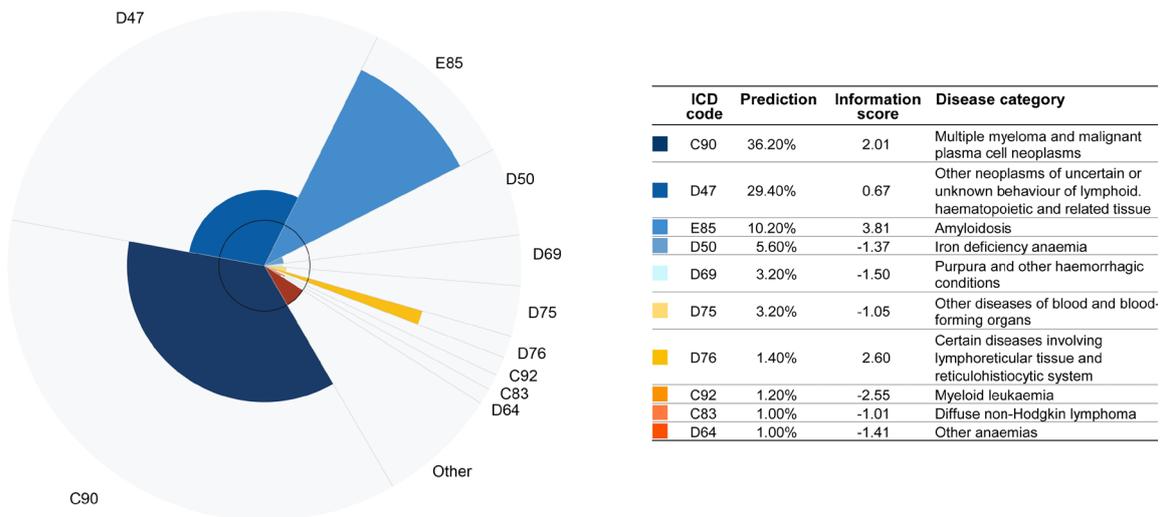

| ICD code | Prediction | Information score | Disease category |
|---|---|---|---|
| C90 | 36.20% | 2.01 | Multiple myeloma and malignant plasma cell neoplasms |
| D47 | 29.40% | 0.67 | Other neoplasms of uncertain or unknown behaviour of lymphoid. haematopoietic and related tissue |
| E85 | 10.20% | 3.81 | Amyloidosis |
| D50 | 5.60% | -1.37 | Iron deficiency anaemia |
| D69 | 3.20% | -1.50 | Purpura and other haemorrhagic conditions |
| D75 | 3.20% | -1.05 | Other diseases of blood and blood-forming organs |
| D76 | 1.40% | 2.60 | Certain diseases involving lymphoreticular tissue and reticulohistiocytic system |
| C92 | 1.20% | -2.55 | Myeloid leukaemia |
| C83 | 1.00% | -1.01 | Diffuse non-Hodgkin lymphoma |
| D64 | 1.00% | -1.41 | Other anaemias |

**Figure 3. Graphical representation of the predictive model results.** Ten most probable diseases are depicted in a polar chart with varying radii. Every chart slice represents a disease, while the angle corresponds to the predicted (posttest) disease probability, whereas its radius is proportional to the logarithm of the ratio between posttest and pretest (prevalence) probability or the information score.

This way of visualizing the results (Fig. 3) emphasizes the most probable diseases, as well as the diseases with the highest information scores. To reach a final diagnosis, a physician should consider both kinds of salient diseases. This is especially useful since predicted probabilities may not be optimally calibrated, as machine learning classifiers often tend to favor the more frequent classes (diseases) and neglect the less frequent ones.

**Data availability**

Both predictive models are available at www.smartbloodanalytics.com upon registration. Twenty clinical test cases with all the data and predicted results can be found in Supplementary S2 Data online. The study was approved by the Slovenian National Medical Ethics Committee (No. 103/11/15).

## Results

**Random forest predictive models**

Utilizing a random forest algorithm, we generated two different predictive models: Smart Blood Analytics Hematology 181 (SBA-HEM181) that was trained with 181 different blood parameters and Smart Blood Analytics Hematology 61 (SBA-HEM061), trained with 61 parameters. Both training sets contained 43 different categories of diseases. We validated both classifiers using stratified tenfold cross-validation. The SBA-HEM181 prediction accuracy was 59%.

Surprisingly SBA-HEM061 performed on a par with the SBA-HEM181 predictive model, attaining a prediction accuracy of 57%. If the first five predictions are taken into account, the SBA-HEM181 and SBA-HEM061 model accuracy were 88% and 86%, respectively (Fig. 4). These results are remarkable given the small data coverage and more importantly, the small subset of data used for predictive model building in regard to the data that is needed for diagnosis.

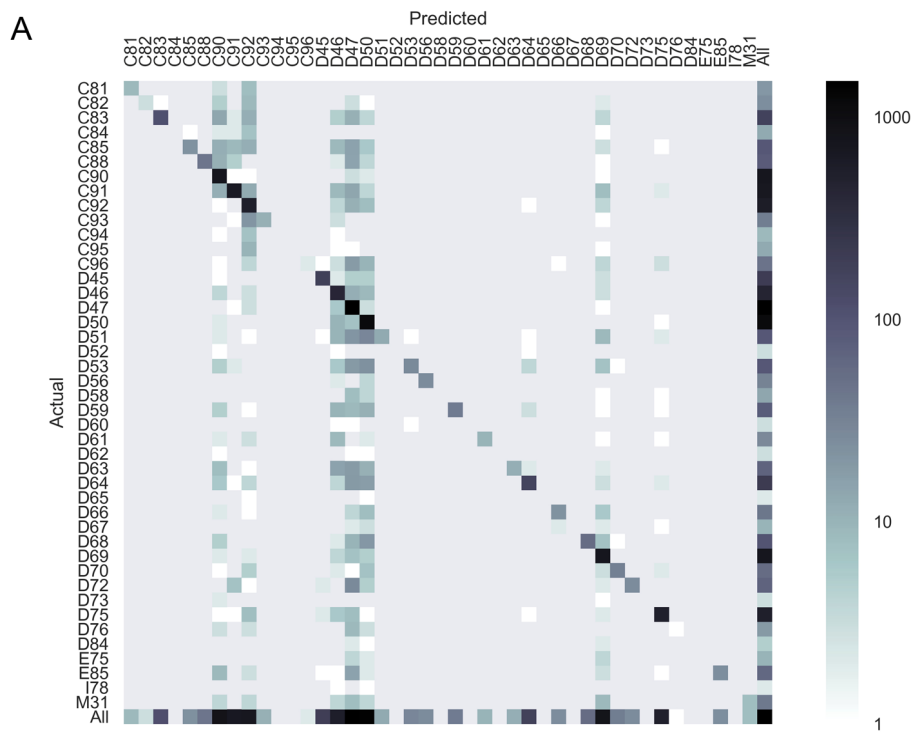

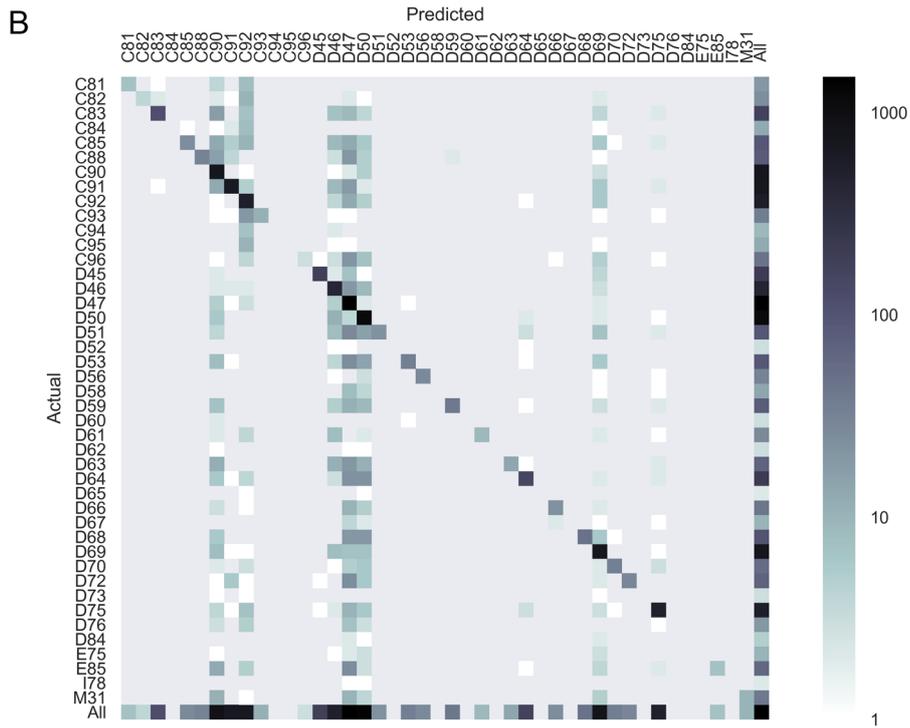

**Figure 4.** Confusion matrix for the five most probable diseases for both predictive models (A) SBA-HEM181 and (B) for SBA-HEM61. Each column of the matrix represents the instances in the predicted diseases, while each row represents the instances in the actual diseases. The frequencies are marked on a logarithmic scale.

Judging from both AUCs and the shape of the ROC curves (Fig. 5) SBA-HEM061 performed slightly better for the less prevalent diseases. Of Interest are the differences between the macro-averaged and micro-averaged ROC curves. The micro-averaged ROC curve lies above the macro-averaged one, which indicates that both predictive models, on average, underperform in the case of rare diseases (Fig. 5).

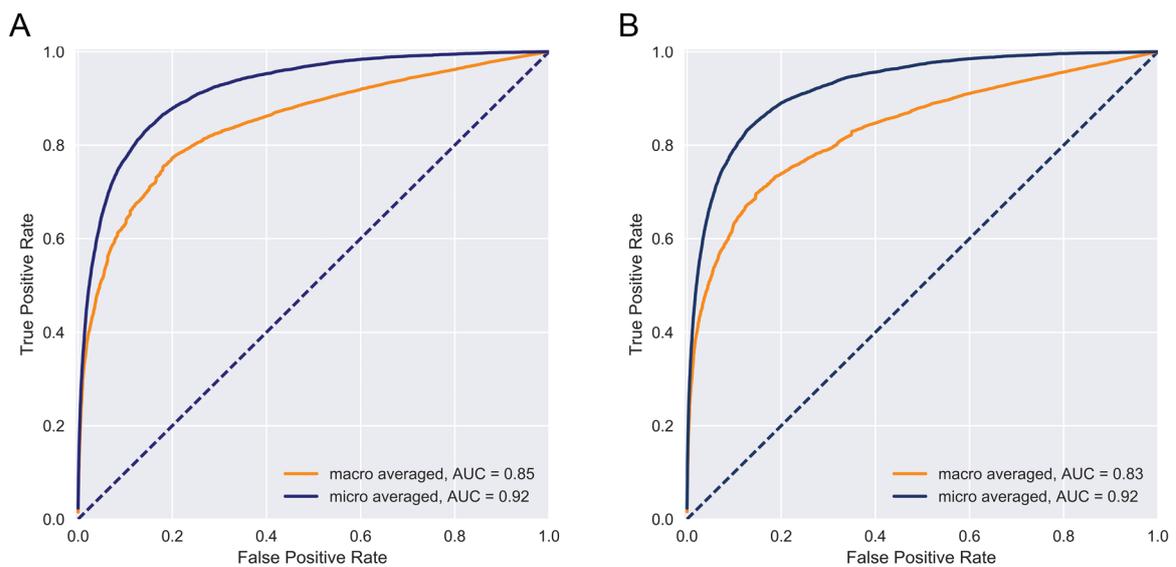

**Figure 5.** Macro- and micro-averaged ROC curves with (a) a full set of 181 parameters and (b) a basic set of 61 parameters. The curves almost overlap while the AUC are almost identical.

The ROC curves for SBA-HEM181 (Fig. 5A) and SBA-HEM061 models (Fig. 5B) are very similar and have practically the same area under the curves (AUC) showing that both models perform equally. We confirmed this hypothesis by applying the Wilcoxon signed rank test[23]. The null hypothesis $H_0$ was that the median ranking of correct predictions for both models are significantly different. Not surprisingly, the p-values obtained on 8233 results using a tenfold cross-validation show that we cannot reject the null hypothesis (p>0.35). We can therefore conclude that the ranking of correct predictions do not differ significantly between the two models, i.e., the reduced model performs as well as the complete one.

**Model versus physician comparison**

To compare the performance of our predictive models with that of a physicians, based only on laboratory blood tests as the input data, we performed a clinical test. Five hematology specialists and 8 specialists of internal medicine were given 20 hematological cases. Our classifier achieved an accuracy of 0.60 (0.55 SBA-HEM61) compared to an accuracy of 0.62 achieved by the hematology specialists (Fig. 6A) and an accuracy of 0.26 achieved by the non-hematology internal medicine specialists (Fig. 6C). When taking into account the first five predicted diseases, accuracy increases to 0.90 for the prediction model SBA-HEM181 (0.85 SBA-HEM61) and is 0.77 for the hematology specialists (Fig. 6B). The Internal medicine specialists only predicted one possible disease.

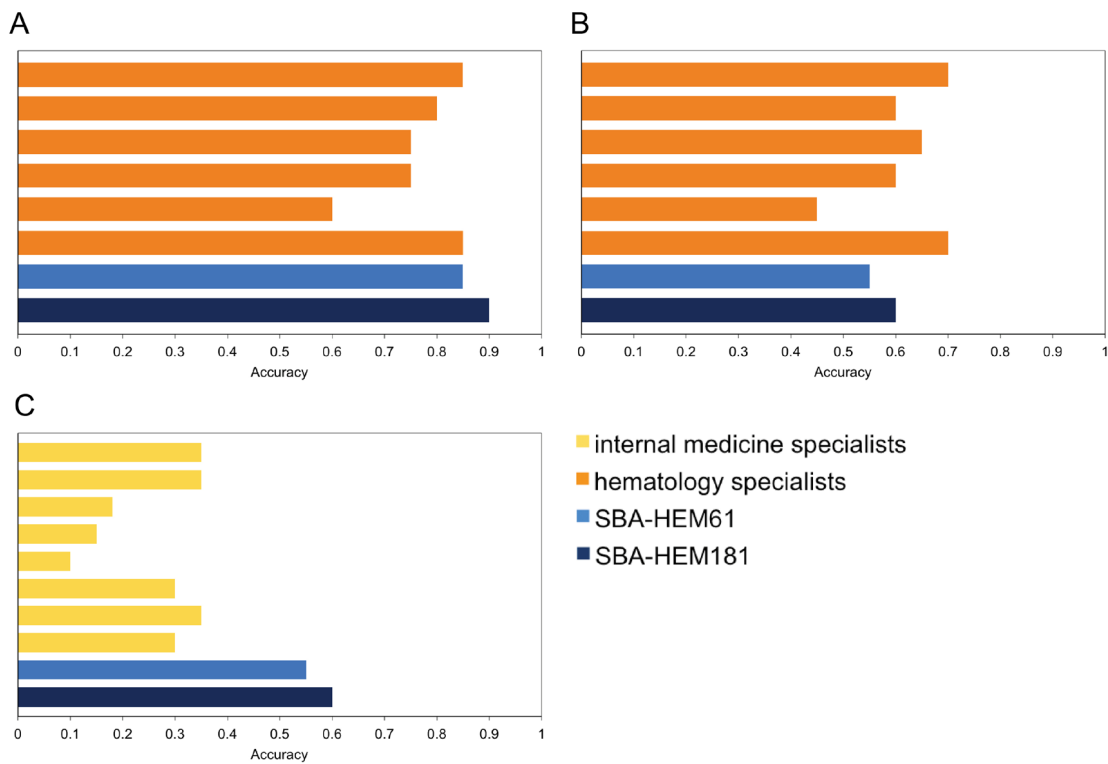

**Figure 6. Comparison of the accuracy of internal medicine specialists with both predictive models** (A) Accuracy of the five hematology specialists compared to both predictive models, when taking in to the account the five most probable predicted diseases. (B) Accuracy of the five hematology specialists compared to both predictive models, when taking in to the account only the most probable predicted disease. (C) Accuracy of the eight non-hematology internal medicine specialists compared to both predictive models, when taking in to account the most probable predicted disease

To compare the physicians' results with the results from our models, we applied the Wilcoxon signed rank tests[23]. The statistical test was performed on a paired sample of 20 results, one set from the physicians, and one from either of the predictive models (SBA-HEM181 or SBA-HEM61). The null hypothesis was that the results (median ranks of correct predictions) of the hematology specialists and that of the models do not differ significantly. For SBA-HEM181 and SBA-HEM61 the Wilcoxon signed-rank test established that there is no significant difference ($p>0.05$) between either of the models and the hematology specialists. When compared with internal medicine specialists, the test result was significant ($p<0.01$) in favor of the predictive models, which perform better (Fig. 3).

**Illustrative example**

A 65-year old man that had been exhibiting diffuse abdominal pain, tiredness, weight loss, back pain and spontaneous bruising over the last 9 months was admitted to the UMCL for hormonal evaluation of an adrenal tumor revealed by computer tomography five months earlier. The standard hormonal evaluation excluded the adrenal tumor's hormonal autonomy, but serum protein evaluation revealed increased serum free light chains indicating a plasma cell disorder. A bone marrow biopsy was performed and the patient was discarded without diagnosis and treatment, while waiting for the bone marrow histology results. After discharge, the laboratory results were analyzed on the Smart Blood Analytics website and SBA-HEM181 model proposed plasmocytoma (ICD category code D90) as the second most probable disease using only admission laboratory results and as the most probable hematological disease using all of the patients laboratory results during hospitalization (Fig. 3). His "normal" laboratory results from two years earlier were analyzed also and the SBA-HEM181 model proposed plasmocytoma as the fourth most likely disease at that time. The patient has since received the hematology specialist's confirmation that he had plasmocytoma and amyloidosis (ICD category code E85) one month after admission.

This example demonstrates how our model could help physicians in facilitating the diagnostic procedure and in the future reduce the number of tests, particularly in insidious diseases, such as plasmocytoma, that begins several years before symptoms appear. It could help patients in the process of requesting a second opinion.

## Discussion

In this study, we show that the machine learning approach, using a random forest algorithm with a large amount of multianalyte sets of hematologic disease laboratory blood test results, is able to interpret the results and predict diseases on a par with an experienced hematology specialist, while outperforming an internal medicine specialist by a margin of more than two. Our study consists of 43 possible disease categories (classes), with a majority class prevalence of 18.4% and a Shannon entropy of 3.95 bit, which is fairly close to the theoretical maximum of 5.42 for 43 classes. This problem is difficult for a machine learning task in the medical domain; according to the UCI repository[24], where more than 50% of medical datasets consist of only two classes, dealing with a single disease (present/absent). In addition, more than 75% of missing attribute (parameter) values do not help either. According to these numbers, a classification (diagnostic) accuracy of 0.57 (SBA-HEM61) and 0.59 (SBA-HEM181) for the most probable disease represents excellent results, comparable to that of a hematology specialist and is far beyond the expected accuracy scores for the simple majority (0.184) or the random predictive model (0.093).

Our predictive models' usefulness in hematological diseases diagnosis was also confirmed in a clinical test in which both predictive models were able to diagnose the type of hematological diseases as well as an experienced hematology specialist and significantly better than a general internal medicine specialists. We were able to show that the accuracy of prediction of SBA-HEM061 and SBA-HEM181, using admission blood laboratory tests, were 0.55 and 0.60, respectively, while hematology specialist's accuracy was 0.60; the three results are not significantly different (Wilcoxon signed-rank test, p>0.05,). However, taking into account the five most probable diagnoses proposed by SBA-HEM181 and the hematology specialists, the accuracy of SBA-HEM181 and that of the hematology specialists improved to 0.90 and 0.77, respectively (significant difference, p<0.05).

Accordingly, SBA could help those physicians not specialized in hematology in facilitating the diagnostic procedure, especially by suggesting proper and early patient's referral. In addition, this study showed that many laboratory tests are not needed to arrive at a correct diagnosis, but may be otherwise useful for additional confirmations. The predictive model SBA-HEM061, trained on only a third of the available laboratory variables (routine laboratory tests that were most frequently used by physicians), was equally successful in predicting the most probable diseases, determined by hematology specialists (57% accuracy). Moreover, the accuracy of SBA-HEM061 for predicting the discharge diagnoses determined by hematology specialists increased to 90%. This suggests there is substantial information redundancy as well as interdependency in laboratory tests, since the accuracy of SBA-HEM061 in predicting diagnoses based on laboratory results was equivalent to using all 181 laboratory tests and the 61 routinely used laboratory tests. From this we can conclude that a machine learning approach in laboratory diagnostics could aid physicians in making early diagnoses of a disease using fewer laboratory tests[9].

The results of this study are encouraging, and oppose established practice and reflections of physicians. In clinical practices, a physician's ability to quickly reach a diagnosis and decide on a management plan is extraordinary and resembles an "art", since much of the process involves skills in clinical decision-making, the essence of which most physicians find hard to describe in an actionable manner. This view is further reinforced by a study showing that a history of a patient, physical examination and laboratory investigation led to a final diagnosis in 76%, 10% and 11% of the cases, respectively[25]. Accordingly, most of physicians are convinced that laboratory test results, viewed in isolation, are typically of a limited diagnostic value in a differential diagnosis, especially in situations where medical knowledge and experience play a significant role[10]. Such an opinion is understandable, considering the absence of published articles on advanced analytical and machine learning approaches in blood laboratory diagnostics. Therefore, it would be useful for physicians to use a machine learning approach in the interpretation of blood laboratory test results. Instead of ordering an increasing number of unnecessary and costly tests, physicians could improve his or her interpretation of the test data by using clinical decision support systems, based on a machine learning approach. These machine learning models could help physicians in the interpretation of multianalyte sets of many individual blood laboratory test results. This is important, since it seems that physicians use even individual laboratory tests insufficiently. In 15% of cases, physicians admitted to not completely understanding what tests they were ordering and 25% admitted to being confused by the results that come back[26]. The sheer volume of tests available and the rapid rate of their increase make a physicians' understanding and proper use of the results to improve the outcome even worse.

Machine learning in blood laboratory tests interpretation could be particularly useful for general practitioners and other physicians not familiar with detailed hematological diagnostics, since our models predict the most probable hematological diseases, using only a patients' blood laboratory results. This can also aid physicians to correctly refer patients. The usefulness of such models would be greatly improved if they were to become an integral part of medical information systems i.e., as decision support for health care providers and spontaneously by including a list of probable diseases in the laboratory report. At the same time, it could stimulate the production of consolidated, large-scale, digitalized databases of patient information that, by itself, could change the course of medicine.

In a time of increasing patient awareness and a desire to be actively involved in treatment, patients and their relatives could use a diagnostic tool like ours, in the process of requesting a second opinion. Thus, user-friendly Smart Blood Analytics website (www.smartbloodanalytics.com) with numerical and graphical representation of proposed diseases was developed and tested prospectively on actual patients. The most interesting example was our illustrative case, in which our model was able to predict the correct disease from laboratory results obtained several years prior to the occurrence of the existing disease and its symptoms.

The excellent results we obtained in our study in the application of machine learning in hematology has encouraged further work within the field of internal medicine as a whole. Models built for other fields of internal medicine show promising results.

## Conclusion

Every disease originates from or causes changes on a cellular and molecular level, which are directly or indirectly almost always detectable in changes in blood parameter values. These changes can be huge and physicians observe them by monitoring the blood parameter values that are out of the normal ranges. However, small changes and/or interactions between multiple different blood parameters that are equally important for detecting pathological patterns (disease "fingerprints"), can be easily overlooked. This implies that the value of blood test results is often underestimated. Furthermore, a machine learning approach is able to recognize disease-related blood laboratory patterns beyond current medical knowledge. This results in higher diagnostic accuracy compared to traditional quantitative interpretation based on reference ranges for blood parameters. Taking a machine learning approach in blood laboratory-based diagnosis could lead to a fundamental change in differential diagnosis and result in the modification of currently accepted guidelines. Nevertheless, physicians will always be needed to interact with patients[4] and should retain access to all raw laboratory data.

In conclusion, a machine learning approach using a random forest algorithm with a sufficiently large data set indicates substantial information redundancy and unobserved potential of laboratory blood test results in diagnosing disease. SBA predictive models show great promise in medical laboratory diagnosis that could be of considerable value to physicians and patients alike, while having widespread beneficial impacts on the cost of healthcare.

        Problems? *Journal of Machine Learning Research* **15**, 3133-3181, doi:10.1016/j.csda.2008.10.033 (2014).

18    Provost, F., Fawcett, T. & Kohavi, R. in *Proceedings of the Fifteenth International Conference on Machine Learning* 445-453 (1997).

19    Fawcett, T. An introduction to {ROC} analysis. *Pattern Recognition Letters* **27**, 861-874, doi:10.1016/j.patrec.2005.10.010 (2006).

20    Sokolova, M. & Lapalme, G. A Systematic Analysis of Performance Measures for Classification Tasks. *Inf. Process. Manage.* **45**, 427-437, doi:10.1016/j.ipm.2009.03.002 (2009).

21    Benish, W. A. Relative Entropy as a Measure of Diagnostic Information. *Medical Decision Making* **19**, 202-206, doi:10.1177/0272989X9901900211 (1999).

22    Oruç, Ö. E. & Kanca, A. Evaluation and Comparison of Diagnostic Test Performance Based on Information Theory. *International Journal of Statistics and Applications* **1**, 10-13 (2011).

23    Conover, W. Statistics of the Kolmogorov-Smirnov type. *Practical nonparametric statistics*, 428-473 (1999).

24    Lichman, M. UCI machine learning repository, 2013. **8** (2015).

25    Peterson, M. C., Holbrook, J. H., Von Hales, D., Smith, N. L. & Staker, L. V. Contributions of the history, physical examination, and laboratory investigation in making medical diagnoses. *The Western journal of medicine* **156**, 163-165, doi:10.1097/00006254-199210000-00013 (1992).

26    Hickner, J. *et al.* Primary Care Physicians' Challenges in Ordering Clinical Laboratory Tests and Interpreting Results. *The Journal of the American Board of Family Medicine* **27**, 268-274, doi:10.3122/jabfm.2014.02.130104 (2014).

## Acknowledgements

We thank hematology specialists at Hematology Department of UMCL: Irena Preložnik Zupan, Biljana Todorova, Barbara Skopec, Matjaž Sever and Jože Pretnar for participating at our clinical test and helpful discussions, List d.o.o. for preparing anonymous medical data and Hardlab d.o.o. for developing intuitive Smart Blood Analytics web interface.
## Author Contributions

Conceptualization, ManN, MarN, MatN, MK, GG; Methodology MK, MarN, GG; Software, MK; Validation, MatN, MK, PČ, MB, GG; Formal Analysis, MK; Investigation, MK, MarN, MatN, GG, PČ, MB; Resources, MarN, MB, PČ; Data Curation, MarN, MatN, GG, MK, MB; Writing, GG, MK, MB, MarN; Supervision, MarN; Funding Acquisition, MarN, MatN.

## Additional Information

Competing financial interests: Smart Blood Analytics Swiss SA (SBA) fully funded this research. MarN is SBA CEO, MK, MatN and GG are SBA advisors.

15